%% file: main.tex
\documentclass[a4paper,5p,times,number]{elsarticle}

\usepackage{amsmath,amssymb,amsfonts}

\usepackage{booktabs}
\usepackage{array}
\usepackage{caption}
\usepackage{placeins}
\usepackage{stfloats}
\usepackage{multirow}
\usepackage{graphicx}
\usepackage{tabularx}
\usepackage{array}
\usepackage{makecell}
\usepackage{ragged2e}
\usepackage{booktabs}
\usepackage{adjustbox}

\newcolumntype{L}[1]{>{\raggedright\arraybackslash}p{#1}}
\newcolumntype{Y}{>{\raggedright\arraybackslash}X}
\newcolumntype{C}[1]{>{\centering\arraybackslash}p{#1}}
\newcolumntype{J}[1]{%
	>{\setlength{\parindent}{0pt}\justifying\arraybackslash}p{#1}%
}

\usepackage{url}

\usepackage[hidelinks]{hyperref}
\usepackage{setspace}
\hypersetup{
  breaklinks=true,
  hypertexnames=false,
  pdftitle={Tissue-Mixture Entropy-Weighted Reconstruction for Partial-Volume-Aware Brain MRI Super-Resolution},
  pdfauthor={Xiao Tong; Wenyun Yang; Ziheng Zhang; Jingzhi Han; Zhaochu Luo; Jinbo Yang},
  pdfkeywords={MRI super-resolution; Partial-volume effect; Tissue-mixture entropy; PVE-balanced reconstruction}
}

\journal{Computer Methods and Programs in Biomedicine}

\begin{document}

\begin{frontmatter}

\title{Tissue-Mixture Entropy-Weighted Reconstruction for Partial-Volume-Aware Brain MRI Super-Resolution}

\author[pku,whio]{Xiao Tong}
\ead{tongxiao@stu.pku.edu.cn}

\author[pku]{Wenyun Yang}
\ead{yangwenyun@pku.edu.cn}

\author[magnvue]{Ziheng Zhang}
\ead{zhangziheng@magnvue.com}

\author[pku]{Jingzhi Han}
\ead{hanjingzhi@pku.edu.cn}

\author[pku]{Zhaochu Luo}
\ead{zhaochu.luo@pku.edu.cn}

\author[pku,whio]{Jinbo Yang\corref{cor}}
\ead{jbyang@pku.edu.cn}

\affiliation[pku]{organization={State Key Laboratory of Artificial
            Microstructure and Mesoscopic Physics, School of Physics,
            Peking University},
            city={Beijing},
            postcode={100871},
            country={P.~R. China}}

\affiliation[whio]{organization={Weihai Institute of Oceanology,
            Peking University},
            city={Weihai},
            postcode={264209},
            country={P.~R. China}}

\affiliation[magnvue]{organization={Beijing MagnVue Medix Co., Ltd.},
            city={Beijing},
            postcode={100085},
            country={P.~R. China}}

\cortext[cor]{Corresponding author. Tel.: +86 10 62753459; Fax: +86 10 62751615.}

\begin{abstract}
\noindent\textbf{Background and Objectives:}
Full-image objectives in brain magnetic resonance imaging (MRI) super-resolution (SR) can underweight tissue-transition regions affected by the partial-volume effect (PVE), as these regions occupy a small fraction of the image. Binary boundaries further provide only a discrete approximation of continuous tissue mixtures within a voxel.

\noindent\textbf{Methods:}
We propose Anatomy-Guided Gaussian-Parameter Warping with PVE-Balanced Reconstruction (AGW-PBR), combining a low-resolution (LR)-only reconstruction backbone with a PVE-aware training objective. The backbone uses LR-derived anatomical guidance, soft latent assignment, and bounded residual warping. Quality-controlled tissue fractions are converted into tissue-mixture entropy to spatially weight reconstruction within validated PVE support. PVE sidecars are used only during training, while inference requires only the LR image. Downstream utility is further evaluated through zero-shot transfer to whole-tumor segmentation on BraTS2023.

\noindent\textbf{Results:}
AGW-PBR improves reconstruction across 2$\times$ and 4$\times$ SR on IXI and achieves the lowest normalized gradient-vector reconstruction error at both CSF--GM and GM--WM interfaces at 4$\times$. Ablation studies verify the contributions of PVE-aware weighting and soft latent assignment. The PVE-free AGW backbone also maintains strong performance on fastMRI. On BraTS2023, AGW-PBR achieves competitive whole-tumor Dice and the lowest HD95 under direct zero-shot transfer.

\noindent\textbf{Conclusions:}
AGW-PBR improves brain MRI SR while preserving tissue-transition information relevant to downstream analysis. The results support tissue-mixture entropy as an effective supervision signal for partial-volume-aware MRI reconstruction.

\end{abstract}

\begin{keyword}
	MRI super-resolution; Partial-volume effect; Tissue-mixture entropy; PVE-balanced reconstruction
\end{keyword}

\end{frontmatter}

\input{sections/introduction}

\input{sections/related_work}
\input{sections/method}
\input{sections/experiment}
\input{sections/conclusion}


\section*{CRediT authorship contribution statement}
\textbf{Xiao Tong:} Conceptualization, Methodology, Software, Validation, Formal analysis, Writing -- original draft. \textbf{Wenyun Yang:} Conceptualization, Supervision, Writing -- review \& editing. \textbf{Ziheng Zhang:} Resources, Data curation. \textbf{Jingzhi Han:} Validation, Investigation. \textbf{Zhaochu Luo:} Supervision, Funding acquisition, Writing -- review \& editing. \textbf{Jinbo Yang:} Supervision, Funding acquisition, Project administration.

\section*{Declaration of generative AI and AI-assisted technologies in the manuscript preparation process}
During the preparation of this article, the authors used ChatGPT to assist with language polishing and grammatical corrections. After using  this tool, the authors reviewed and edited the manuscript as needed and take full responsibility for the content of the published article.

\section*{Declaration of competing interest}
Ziheng Zhang is employed by Beijing MagnVue Medix Co., Ltd. The remaining authors declare that they have no competing financial interests or personal relationships that could have appeared to influence the work reported in this paper.

\section*{Data availability}
The IXI, fastMRI and BraTS2023 datasets used in this study are publicly available from their respective repositories, subject to the applicable data-use conditions. The authors are not authorized to redistribute the original data.

The subject-level split manifests, preprocessing and quality-control scripts, model configurations, evaluation code, and derived subject-level results are available from the corresponding author upon reasonable request.

\section*{Acknowledgements}
This work was supported by the National Key Research and Development Program of China (Grant No.~2023YFB3507000) and the Research Fund of Weihai Institute of Oceanology, Peking University.

\bibliographystyle{elsarticle-num}
\bibliography{references}

\end{document}

%% file: sections/introduction.tex
\section{Introduction}
\label{sec:introduction}

Magnetic resonance imaging (MRI) provides strong soft-tissue contrast for structural assessment and quantitative image analysis. High spatial resolution is particularly important in brain MRI for delineating cortical folds, tissue interfaces, and small anatomical structures. However acquiring images at a higher spatial resolution generally requires longer scan time and introduces trade-offs in signal-to-noise ratio, motion sensitivity, and clinical throughput~\citep{arssr,downsample}. Image super-resolution (SR) provides a computational alternative by estimating a high-resolution (HR) image from a low-resolution (LR) observation without modifying scanner hardware.

Brain MRI SR is particularly sensitive to tissue transitions. At finite spatial resolution, voxels near cerebrospinal fluid (CSF), gray matter (GM), and white matter (WM) interfaces can contain contributions from multiple tissues, producing partial-volume effects (PVEs)~\citep{ballester2002estimation,van2003unifying,tohka2004fast}. These regions occupy a small fraction of the image, so their reconstruction errors can receive limited aggregate influence under full-image objectives dominated by homogeneous tissue. Boundary-aware losses emphasize contours~\citep{liu2019edge}, while tissue fractions provide a continuous description of within-voxel composition. Existing PVE studies primarily address segmentation and tissue-fraction estimation~\citep{van2003unifying,tohka2004fast,zhang2001fast}.

We propose Anatomy-Guided Gaussian-Parameter Warping with PVE-Balanced Reconstruction (AGW-PBR), which consists of an LR-only reconstruction backbone and a PVE-balanced training objective. The reconstruction backbone uses LR-derived structural guidance, compact soft latent assignment, and bounded grid-anchored residual warping. During training, quality-controlled CSF/GM/WM pseudo-label sidecars provide tissue-mixture entropy, which defines mean-normalized spatial reconstruction weights within validated PVE support. The sidecars remain fixed training resources. Here, ``Gaussian-parameter'' refers to the displacement and gating parameterization of the warping module.

The contributions of this work are summarized as follows:
\begin{itemize}
	\item We introduce PVE-Balanced Reconstruction, which uses mean-normalized tissue-mixture entropy to increase the contribution of partial-volume tissue transitions during training while retaining full-image reconstruction supervision.
	\item We develop an LR-only Anatomy-Guided Gaussian-Parameter Warping (AGW) reconstruction framework based on image-derived structural guidance, compact soft latent assignment, and bounded grid-anchored residual warping.
	\item We establish quality-controlled PVE sidecars and independent SynthSeg-based evaluation of regional fidelity and CSF--GM/GM--WM tissue-transition gradients. Boundary-profile RMSE provides complementary analysis on HR-eligible profiles.
	\item We evaluate the AGW-PBR model on the IXI dataset at $2\times$ and $4\times$ with subject-level paired analysis and further assess downstream information retention through zero-shot transfer to a frozen BraTS2023 whole-tumor segmentation probe. We separately evaluate the AGW reconstruction backbone on the fastMRI dataset under an independent SR-only protocol without PVE sidecars.
\end{itemize}

%% file: sections/related_work.tex
\section{Related Work}
\label{sec:related_work}

\subsection{MRI Super-Resolution Architectures}

MRI super-resolution (SR) reconstructs high-resolution (HR) images or volumes from low-resolution (LR) observations. Early methods relied on interpolation, projection-based reconstruction, and handcrafted priors based on smoothness, consistency, or sparsity~\citep{keys1981,stark1989,sun2008image,yang2010image}. Learning-based methods primarily differ in their modeling of spatial context: convolutional networks extract local and multiscale features, while Transformer-based models capture long-range dependencies through attention~\citep{fmisr,rcab,swinir,forigua2022}.

Continuous representations enable arbitrary-scale reconstruction by predicting intensities at queried coordinates. General SR methods such as LIIF, LTE, and CiaoSR use coordinate-conditioned features~\citep{liif,lte,ciaosr}. For MRI, ArSSR extends implicit representations to arbitrary-scale 3-D reconstruction, while SA-INR introduces locally aware spatial attention for flexible inter-slice upsampling~\citep{arssr,wang_sainr_2024}. NExpR and GaussianSR further improve computational efficiency or scale flexibility~\citep{nexpr2025,gaussiansr}.

Diffusion-based SR reconstructs HR images through conditional generative refinement, with later methods reducing sampling cost through compact priors, residual-shifting trajectories, and continuous-scale formulations~\citep{sr3,diffir,resshift,idm}. In MRI SR, Res-SRDiff adopts residual-shifting diffusion, Partial Diffusion Models initialize reverse diffusion from an LR-derived intermediate latent, and PRDDiff progressively restores high-frequency $k$-space components for arbitrary-scale in-plane reconstruction~\citep{ressrdiff,zhao_partial_diffusion_2025,wang_prddiff_2026}.

These architectures improve representation capacity, scale flexibility, and high-frequency reconstruction. Their reconstruction objectives are generally aggregated over the image, leaving the relative contribution of partial-volume tissue transitions largely implicit.

\subsection{Anatomical, Segmentation, and Tissue-Aware Guidance}

Anatomical guidance can be introduced through an additional MR contrast. Progressive fusion, cross-modality Transformers, and deformable attention integrate or align complementary contrast information across reconstruction stages~\citep{downsample,minet,fang2022crossmodality,zou2023multiscale,dance2025}. Such guidance provides anatomical correspondence but requires additional acquisition and registration.

Anatomical priors can also be embedded in learned representations or training objectives. PGASR conditions a discrete anatomical codebook on rough structure maps and aligns representations across slices and LR/HR domains, while SCSR constrains SR training with a frozen cortical-ribbon segmentation model~\citep{luo_pgasr_2025,wu_scsr_2024}. BME-X instead predicts voxel-wise tissue classes (background, CSF, GM, and WM) and concatenates these probabilities with the input image for reconstruction~\citep{sun_bmex_2025}. These categorical tissue cues differ from continuous tissue fractions, which represent mixed tissue composition within a voxel.

\subsection{Partial-Volume Effects and Tissue-Mixture Modeling}

Partial-volume effects (PVEs) arise when finite spatial resolution causes a voxel to contain signals from multiple tissues, commonly at CSF/GM/WM interfaces in brain MRI~\citep{ballester2002estimation,tohka2004fast}. Prior work has mainly addressed PVE through mixture modeling, segmentation, and continuous tissue-fraction estimation using spatial and statistical models~\citep{van2003unifying,tohka2004fast,zhang2001fast}.

Boundary information and tissue fractions describe complementary properties of an interface. Boundary-aware objectives can increase the contribution of reconstruction errors near structural contours~\citep{liu2019edge}. A boundary marks where a transition occurs, whereas a CSF/GM/WM fraction vector describes the relative tissue composition within a voxel. Tissue fractions therefore provide a continuous description of tissue mixture that is not captured by a binary contour alone.

Our study uses quality-controlled CSF/GM/WM PVE pseudo-label sidecars to derive tissue-mixture entropy for training-time spatial reconstruction weighting, with independent SynthSeg segmentations used for regional evaluation.

%% file: sections/method.tex
\section{Methods}
\label{sec:methods}

\subsection{Problem Formulation and Method Overview}

Let $\Omega=\{1,\ldots,H\}\times\{1,\ldots,W\}$ denote the target Cartesian grid. The scalar images $I_{\mathrm{LR}},I_{\mathrm{HR}},I_{\mathrm{SR}}:\Omega\rightarrow\mathbb{R}$ denote, respectively, the LR-derived observation represented on this grid, its HR target, and the reconstructed output. With a fixed degradation-and-resampling operator $\mathcal{D}$ and a reconstruction network $\mathcal{F}_{\phi}$ parameterized by $\phi$, the problem is
\begin{equation}
\begin{aligned}
	I_{\mathrm{LR}} &= \mathcal{D}(I_{\mathrm{HR}}),
	& I_{\mathrm{SR}} &= \mathcal{F}_{\phi}(I_{\mathrm{LR}}).
\end{aligned}
\label{eq:problem_mapping}
\end{equation}

\begin{figure*}[t]
	\centering
	\includegraphics[width=\textwidth]{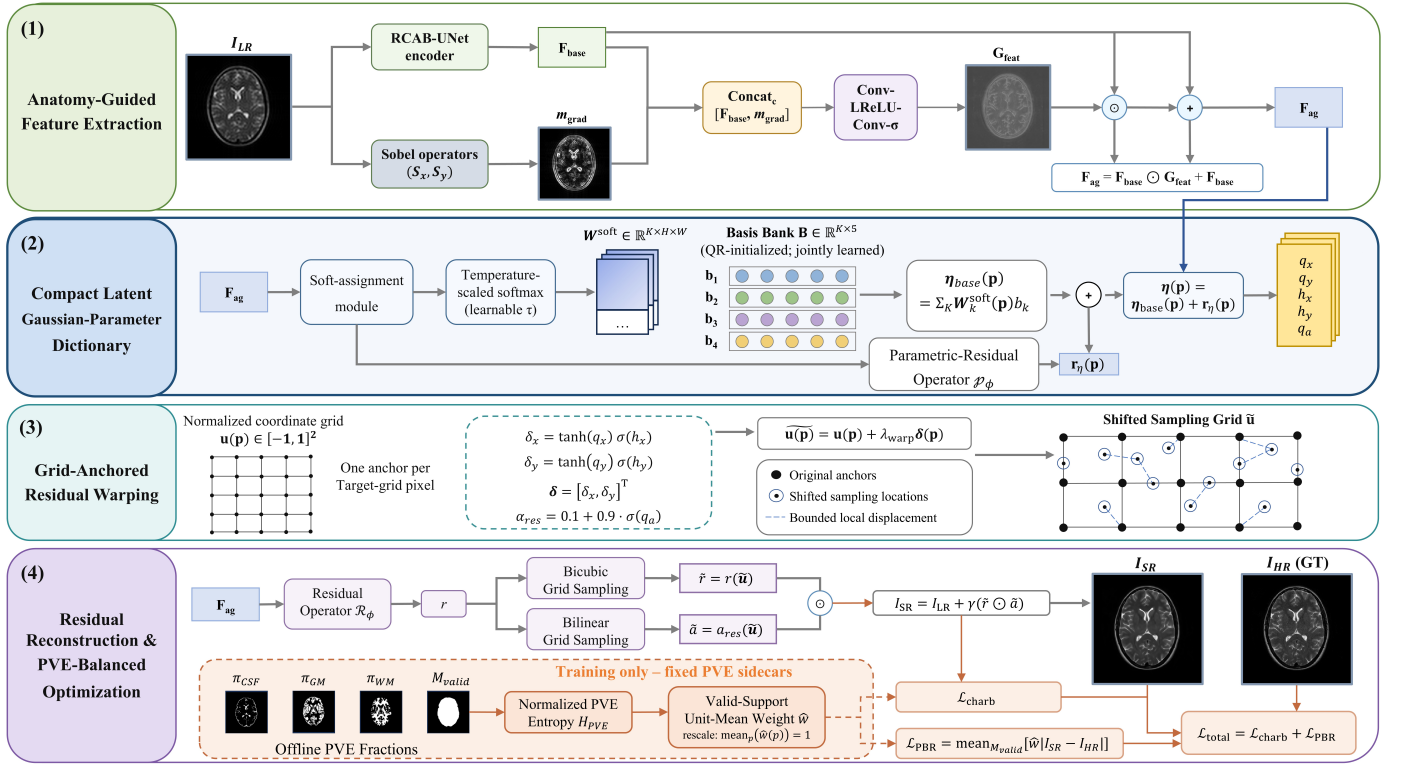}
	\captionsetup{
		justification=justified,
		singlelinecheck=false
	}
	\caption{
		Overview of AGW-PBR, which reconstructs high-resolution MRI in four stages.
		(1) \textbf{Anatomy-Guided Feature Extraction}: an RCAB U-Net extracts base features, while Sobel gradients guide feature modulation to obtain $\mathbf{F}_{\mathrm{ag}}$.
		(2) \textbf{Compact Latent Gaussian-Parameter Dictionary}: temperature-scaled soft assignment combines four anonymous learned bases, and a parameter-residual branch refines the resulting five-dimensional local code.
		(3) \textbf{Grid-Anchored Residual Warping}: the local code is decoded into bounded displacements and a residual gate, which define the shifted sampling grid $\widetilde{\mathbf{u}}$.
		(4) \textbf{Residual Reconstruction \& PVE-Balanced Optimization}: the residual field and gate are sampled on $\widetilde{\mathbf{u}}$ and combined with $I_{\mathrm{LR}}$ to produce $I_{\mathrm{SR}}$.
		During training, fixed CSF, GM, and WM PVE fractions provide an entropy-derived unit-mean weight for $\mathcal{L}_{\mathrm{PBR}}$, which is combined with $\mathcal{L}_{\mathrm{charb}}$.
	}
	\label{fig:agw_pbr_framework}
\end{figure*}

Anatomy-Guided Gaussian-Parameter Warping with PVE-Balanced Reconstruction (AGW-PBR, Fig.~\ref{fig:agw_pbr_framework}) separates an LR-only reconstruction pathway from a training-only PVE-balanced objective. The reconstruction pathway derives a Sobel-gradient cue from $I_{\mathrm{LR}}$, uses it to modulate learned image features, and predicts a five-dimensional local code through a softly assigned latent basis bank. This code produces bounded displacements and gates for grid-anchored residual warping. During training, fixed offline PVE sidecars weight reconstruction errors within quality-controlled support but never enter the network.

\subsection{Anatomy-Guided Feature Extraction}

A learned RCAB U-Net-style encoder $\mathcal{E}_{\phi}$ maps the LR-derived observation to the base feature tensor
\begin{equation}
	\mathbf{F}_{\mathrm{base}}
	=
	\mathcal{E}_{\phi}(I_{\mathrm{LR}})
	\in\mathbb{R}^{C_f\times H\times W},
\label{eq:base_features}
\end{equation}
where $C_f$ is the feature-channel count. Fixed horizontal and vertical Sobel kernels $\mathbf{S}_x,\mathbf{S}_y\in\mathbb{R}^{3\times3}$ produce scalar response fields and their gradient magnitude:
\begin{equation}
\begin{aligned}
	g_x &= I_{\mathrm{LR}}\ast\mathbf{S}_x,
	& g_y &= I_{\mathrm{LR}}\ast\mathbf{S}_y,\\
	m_{\mathrm{grad}}
	&=\sqrt{g_x^2+g_y^2+\epsilon_g}.
\end{aligned}
\label{eq:sobel_guidance}
\end{equation}
The symbol $\ast$ is reserved for two-dimensional convolution, and $\epsilon_g=10^{-6}$ is a numerical stabilizer. 

The gate predictor is decomposed into explicit intermediate tensors:
\begin{equation}
\begin{aligned}
	\mathbf{F}_{\mathrm{cat}}
	&=\operatorname{Concat}_{c}
	  (\mathbf{F}_{\mathrm{base}},m_{\mathrm{grad}}),\\
	\mathbf{F}_{\mathrm{gate}}
	&=\delta\bigl(\mathcal{C}_1(\mathbf{F}_{\mathrm{cat}})\bigr),\\
	\mathbf{G}_{\mathrm{feat}}
	&=\sigma\bigl(\mathcal{C}_2(\mathbf{F}_{\mathrm{gate}})\bigr),
\end{aligned}
\label{eq:feature_gate}
\end{equation}
where $\operatorname{Concat}_{c}$ denotes channel-wise concatenation, $\mathcal{C}_1$ and $\mathcal{C}_2$ are learned convolutional operators, $\delta$ is a Leaky ReLU, and $\sigma$ is the sigmoid function. Thus, $\mathbf{F}_{\mathrm{cat}}\in\mathbb{R}^{(C_f+1)\times H\times W}$ and $\mathbf{G}_{\mathrm{feat}}\in(0,1)^{C_f\times H\times W}$. Element-wise residual modulation gives
\begin{equation}
	\mathbf{F}_{\mathrm{ag}}
	=
	\mathbf{F}_{\mathrm{base}}
	+
	\mathbf{F}_{\mathrm{base}}\odot\mathbf{G}_{\mathrm{feat}},
\label{eq:anatomy_guided_features}
\end{equation}
where $\odot$ denotes element-wise multiplication between shape-compatible fields or tensors. This image-derived guidance pathway is active during both training and inference.

\subsection{Compact Latent Gaussian-Parameter Dictionary}

An assignment operator $\mathcal{A}_{\phi}$ predicts $K$ basis logits from the anatomy-guided features:
\begin{equation}
	\mathbf{Z}
	=
	\mathcal{A}_{\phi}(\mathbf{F}_{\mathrm{ag}})
	\in\mathbb{R}^{K\times H\times W}.
\label{eq:assignment_logits}
\end{equation}
At each anchor $\mathbf{p}\in\Omega$, the final model normalizes these logits along the $K$-basis axis:
\begin{equation}
	W_k^{\mathrm{soft}}(\mathbf{p})
	=
	\frac{\exp\!\left(Z_k(\mathbf{p})/\tau\right)}
	{\displaystyle\sum_{j=1}^{K}
	 \exp\!\left(Z_j(\mathbf{p})/\tau\right)},
	\quad k=1,\ldots,K.
\label{eq:soft_assignment}
\end{equation}
The temperature $\tau$ is a learnable scalar initialized to $1$. The assignments satisfy $W_k^{\mathrm{soft}}(\mathbf{p})\geq0$ and $\sum_{k=1}^{K}W_k^{\mathrm{soft}}(\mathbf{p})=1$.

Let $\mathbf{B}\in\mathbb{R}^{K\times D_{\eta}}$ be the learned basis bank and let $\mathbf{b}_k\in\mathbb{R}^{D_{\eta}}$ be its $k$-th row. A parameter-residual operator $\mathcal{P}_{\phi}$ predicts $\mathbf{R}_{\eta}\in\mathbb{R}^{D_{\eta}\times H\times W}$. The basis mixture and final local code are
\begin{equation}
\begin{aligned}
	\boldsymbol{\eta}_{\mathrm{base}}(\mathbf{p})
	&=
	\sum_{k=1}^{K}
	W_k^{\mathrm{soft}}(\mathbf{p})\,\mathbf{b}_k,\\
	\mathbf{R}_{\eta}
	&=
	\mathcal{P}_{\phi}(\mathbf{F}_{\mathrm{ag}}),\\
	\boldsymbol{\eta}(\mathbf{p})
	&=
	\boldsymbol{\eta}_{\mathrm{base}}(\mathbf{p})
	+
	\mathbf{r}_{\eta}(\mathbf{p}),
\end{aligned}
\label{eq:latent_parameter_code}
\end{equation}
where $\mathbf{r}_{\eta}(\mathbf{p})$ is the $D_{\eta}$-vector at $\mathbf{p}$ in $\mathbf{R}_{\eta}$. AGW-PBR uses $K=4$ and $D_{\eta}=5$, with
\begin{equation}
	\boldsymbol{\eta}(\mathbf{p})
	=
	[q_x(\mathbf{p}),q_y(\mathbf{p}),
	  h_x(\mathbf{p}),h_y(\mathbf{p}),
	  q_a(\mathbf{p})]^{\mathsf T}.
\label{eq:parameter_decomposition}
\end{equation}
Here, $q_x$ and $q_y$ are raw displacement logits, $h_x$ and $h_y$ are raw displacement-amplitude gate logits, and $q_a$ is the raw residual-intensity gate logit. The basis bank is QR-initialized and jointly optimized with the network. 

\subsection{Grid-Anchored Residual Warping and Reconstruction}

A learned residual operator $\mathcal{R}_{\phi}$ predicts a single-channel scalar field
\begin{equation}
	r
	=
	\mathcal{R}_{\phi}(\mathbf{F}_{\mathrm{ag}})
	\in\mathbb{R}^{H\times W}.
\label{eq:residual_field}
\end{equation}
Five raw components in Eq.~\eqref{eq:parameter_decomposition} are converted into a bounded displacement vector and a residual-intensity gate:
\begin{equation}
\begin{aligned}
	\delta_x(\mathbf{p})
	&=\tanh(q_x(\mathbf{p}))\,\sigma(h_x(\mathbf{p})),\\
	\delta_y(\mathbf{p})
	&=\tanh(q_y(\mathbf{p}))\,\sigma(h_y(\mathbf{p})),\\
	\boldsymbol{\delta}(\mathbf{p})
	&=[\delta_x(\mathbf{p}),\delta_y(\mathbf{p})]^{\mathsf T},\\
	a_{\mathrm{res}}(\mathbf{p})
	&=0.1+0.9\,\sigma(q_a(\mathbf{p})).
\end{aligned}
\label{eq:bounded_displacement_gate}
\end{equation}
Thus, $h_x$ and $h_y$ modulate the displacement amplitude, and $a_{\mathrm{res}}$ gates the predicted residual.

Let $\mathbf{u}(\mathbf{p})\in[-1,1]^2$ be the normalized coordinate of anchor $\mathbf{p}$. The shifted sampling coordinate is
\begin{equation}
	\widetilde{\mathbf{u}}(\mathbf{p})
	=
	\mathbf{u}(\mathbf{p})
	+
	\lambda_{\mathrm{warp}}\,\boldsymbol{\delta}(\mathbf{p}),
\label{eq:shifted_grid}
\end{equation}
where $\lambda_{\mathrm{warp}}$ is a fixed coefficient in normalized-grid coordinates ($\lambda_{\mathrm{warp}}=0.05$). Define $\operatorname{GridSample}_{\kappa}(X,\mathbf{v})$ as sampling field $X$ at normalized coordinates $\mathbf{v}$ using interpolation mode $\kappa$. The warped residual and gate are
\begin{equation}
\begin{aligned}
	\widetilde r(\mathbf{p})
	&=
	\operatorname{GridSample}_{\mathrm{bic}}
	(r,\widetilde{\mathbf{u}}(\mathbf{p})),\\
	\widetilde a(\mathbf{p})
	&=
	\operatorname{GridSample}_{\mathrm{lin}}
	(a_{\mathrm{res}},\widetilde{\mathbf{u}}(\mathbf{p})).
\end{aligned}
\label{eq:grid_sampling}
\end{equation}
The residual sampler is bicubic, the gate sampler is bilinear, and both use reflection padding with aligned corner pixels. The reconstructed image is
\begin{equation}
	I_{\mathrm{SR}}(\mathbf{p})
	=
	I_{\mathrm{LR}}(\mathbf{p})
	+
	\gamma\,
	(\widetilde r\odot\widetilde a)(\mathbf{p}),
\label{eq:residual_reconstruction}
\end{equation}
where $\gamma$ is the fixed residual coefficient ($\gamma=0.1$). Eq.~\eqref{eq:residual_reconstruction} combines one shifted residual sample with each grid anchor.

\subsection{PVE-Balanced Reconstruction Objective}
\label{sec:loss}
Inference requires only $I_{\mathrm{LR}}$; Sobel guidance and target-grid coordinates are generated internally. PVE sidecars are training-only resources, and SynthSeg interfaces are used only for offline evaluation.

For image pair $n$, the tissue-fraction vector at $\mathbf{p}$ is
\begin{equation}
	\boldsymbol{\pi}^{(n)}(\mathbf{p})
	=
	[\pi_{\mathrm{CSF}}^{(n)}(\mathbf{p}),
	 \pi_{\mathrm{GM}}^{(n)}(\mathbf{p}),
	 \pi_{\mathrm{WM}}^{(n)}(\mathbf{p})]^{\mathsf T},
\label{eq:pve_fraction_vector}
\end{equation}
from which the normalized entropy sidecar is constructed as
\begin{equation}
	\begin{aligned}
	H_{\mathrm{PVE}}^{(n)}(\mathbf{p})
	&=
	-\frac{1}{\log 3}
	\sum_{c\in\{\mathrm{CSF},\mathrm{GM},\mathrm{WM}\}}
	\widetilde{\pi}_c^{(n)}(\mathbf{p})
	\log\!\left(
	\widetilde{\pi}_c^{(n)}(\mathbf{p})+\epsilon_h
	\right),\\
	\widetilde{\pi}_c^{(n)}(\mathbf{p})
	&=
	\operatorname{clip}\!\left(\pi_c^{(n)}(\mathbf{p}),0,1\right).
	\end{aligned}
\label{eq:pve_entropy}
\end{equation}
where $\epsilon_h=10^{-8}$; the resulting entropy is then clipped to $[0,1]$. The resulting entropy field and valid-support mask remain fixed during training.

For a current minibatch of $N$ pairs, define the active support
\begin{equation}
	\mathcal{V}_{\mathrm{mb}}
	=
	\{(n,\mathbf{p}):
	  n\in\{1,\ldots,N\},\ 
	  \mathbf{p}\in\Omega,\ 
	  M_{\mathrm{valid}}^{(n)}(\mathbf{p})=1\}.
\label{eq:active_pve_support}
\end{equation}
When $\mathcal{V}_{\mathrm{mb}}$ is nonempty, let $H_{\min}$ and $H_{\max}$ be the minimum and maximum of $H_{\mathrm{PVE}}$ over this joint minibatch support. For $(n,\mathbf{p})\in\mathcal{V}_{\mathrm{mb}}$, the normalized guide is
\begin{equation}
	G^{(n)}(\mathbf{p})
	=
	\begin{cases}
	\dfrac{H_{\mathrm{PVE}}^{(n)}(\mathbf{p})-H_{\min}}
	      {H_{\max}-H_{\min}},
	& H_{\max}-H_{\min}>\epsilon_h,\\[6pt]
	0,
	& H_{\max}-H_{\min}\leq\epsilon_h,
	\end{cases}
\label{eq:pve_guide_normalization}
\end{equation}
Thus, extrema are shared across all valid pixels in the current minibatch, and a constant guide becomes zero. If $\mathcal{V}_{\mathrm{mb}}$ is empty, the implementation returns a safe zero PBR contribution.

For nonempty $\mathcal{V}_{\mathrm{mb}}$, the raw weight, its valid-support mean, and the unit-mean weight are
\begin{equation}
\begin{aligned}
	w_{\mathrm{raw}}^{(n)}(\mathbf{p})
	&=1+\alpha_{\mathrm{pve}}G^{(n)}(\mathbf{p}),\\
	\mu_w
	&=\frac{1}{|\mathcal{V}_{\mathrm{mb}}|}
	  \sum_{(m,\mathbf{q})\in\mathcal{V}_{\mathrm{mb}}}
	  w_{\mathrm{raw}}^{(m)}(\mathbf{q}),\\
	\widehat w^{(n)}(\mathbf{p})
	&=\frac{w_{\mathrm{raw}}^{(n)}(\mathbf{p})}{\mu_w}.
\end{aligned}
\label{eq:pve_weights}
\end{equation}
Here, $\alpha_{\mathrm{pve}}$ controls the relative emphasis on high-entropy locations. By construction, $|\mathcal{V}_{\mathrm{mb}}|^{-1}\sum_{\mathcal{V}_{\mathrm{mb}}}\widehat w=1$.

The PBR term is the valid-support weighted $L_1$ error, including the empty-support safeguard:
\begin{equation}
	\mathcal{L}_{\mathrm{PBR}}
	=
	\begin{cases}
	\dfrac{1}{|\mathcal{V}_{\mathrm{mb}}|}
	\displaystyle\sum_{(n,\mathbf{p})\in\mathcal{V}_{\mathrm{mb}}}
	\widehat w^{(n)}(\mathbf{p})
	\left|I_{\mathrm{SR}}^{(n)}(\mathbf{p})
	      -I_{\mathrm{HR}}^{(n)}(\mathbf{p})\right|,
	& |\mathcal{V}_{\mathrm{mb}}|>0,\\[8pt]
	0,
	& |\mathcal{V}_{\mathrm{mb}}|=0.
	\end{cases}
\label{eq:pbr_loss}
\end{equation}
The global single-channel Charbonnier loss reduces over every image and target-grid location:
\begin{equation}
	\mathcal{L}_{\mathrm{charb}}
	=
	\frac{1}{N|\Omega|}
	\sum_{n=1}^{N}\sum_{\mathbf{p}\in\Omega}
	\sqrt{
	\left(I_{\mathrm{SR}}^{(n)}(\mathbf{p})
	      -I_{\mathrm{HR}}^{(n)}(\mathbf{p})\right)^2
	+\epsilon_c^2},
\label{eq:charbonnier_loss}
\end{equation}
where $\epsilon_c=10^{-3}$. The final training objective is
\begin{equation}
	\mathcal{L}_{\mathrm{total}}
	=
	\mathcal{L}_{\mathrm{charb}}
	+
	\mathcal{L}_{\mathrm{PBR}}.
\label{eq:total_loss}
\end{equation}
Gradients from $\mathcal{L}_{\mathrm{PBR}}$ propagate only through the reconstruction network.

\subsection{Inference Boundary}

At inference, the model uses only $I_{\mathrm{LR}}$, with Sobel guidance and target-grid coordinates generated internally. PVE sidecars are used only for IXI training and pseudo-label quality control, whereas SynthSeg-derived interfaces are used solely to construct offline test-set evaluation masks.

%% file: sections/experiment.tex
\section{Experiments}
\label{sec:experiments}
\captionsetup{format=plain,justification=justified,singlelinecheck=false}
We evaluate AGW-PBR on IXI T$_2$ and the PVE-free AGW backbone on fastMRI T$_2$. IXI provides the primary reconstruction and tissue-interface evaluation, while fastMRI provides an independent SR-only external test.

\subsection{Experimental Settings}
\label{sec:experimental_settings}

\noindent\textbf{Datasets.}
The IXI dataset\footnote{\url{https://brain-development.org/ixi-dataset/}} serves as the primary brain MRI SR benchmark, with 329 training, 47 validation, and 99 test subjects after quality control. All selected T$_2$-weighted slices have a fixed in-plane resolution of $256 \times 256$. We evaluate the AGW-backbone on the fastMRI dataset~\footnote{\url{https://fastmri.med.nyu.edu/}} \citep{knoll2020fastmri}, using 200 subjects (3,200 slices) for training, 30 subjects (480 slices) for validation, and 43 subjects (688 slices) for testing. Table~\ref{tab:dataset_training_budget} summarizes the dataset and training configurations.

\begin{table*}[!tbp]
	\centering
	\caption{Dataset and training-budget summary for the main experiments.}
	\label{tab:dataset_training_budget}
	\begingroup
	\small
	\renewcommand{\arraystretch}{1.18}
	\setlength{\tabcolsep}{0pt}
	\resizebox{\textwidth}{!}{%
	\begin{tabular}{@{}L{0.11\textwidth}@{\hspace{0.80em}}L{0.10\textwidth}@{\hspace{0.80em}}L{0.09\textwidth}@{\hspace{0.80em}}L{0.22\textwidth}@{\hspace{0.80em}}p{0.20\textwidth}@{\hspace{0.80em}}p{0.21\textwidth}@{}}
		\toprule
		Dataset & Slice size & Scale(s) & Split & Training budget & Evaluation scope \\
		\midrule
		IXI (T$_2$) & $256 \times 256$ & 2$\times$, 4$\times$ & 329 train / 47 validation / 99 test subjects & 80 epochs per scale / seed; seeds 42, 43, 44 & Full-image and SynthSeg regional PSNR/SSIM.  \\
		fastMRI (T$_2$) & $320 \times 320$ & 4$\times$ & 200 train / 30 validation / 43 test subjects  & 80 epochs per seed; seeds 42, 43, 44 & Full-image PSNR/SSIM. \\
	\bottomrule
	\end{tabular}
	}
	\endgroup
\end{table*}

\begin{table*}[!tbp]
	\centering
	\caption{Quality-control summary of the IXI PVE pseudo-labels and independent interface masks.}
	\label{tab:pve_certification_summary}
	\begingroup
	\small
	\renewcommand{\arraystretch}{1.16}
	\setlength{\tabcolsep}{0pt}
	\begin{tabular}{@{}L{0.22\textwidth}@{\hspace{1.00em}}L{0.25\textwidth}@{\hspace{1.50em}}p{0.46\textwidth}@{}}
		\toprule
		Item & Value & Note \\
		\midrule
		Retained cohort
		& 475 subjects
		& 329 training, 47 validation, and 99 test subjects pass brain-mask, registration, tissue-fraction, and sidecar quality control. \\
		
		Validated sidecars
		& 43,141 slices
		& Includes 8,998 test slices with unique subject--slice correspondence to the T$_2$ images. \\
		
		Multimodal preprocessing
		& T$_1$ / T$_2$ / PD
		& T$_1$ and PD are aligned to T$_2$, and the same T$_2$-derived HD-BET mask is applied to all three modalities before FAST. \\
		
		Tissue fractions
		& CSF / GM / WM
		& Multi-channel FAST estimates the three partial-volume fractions on the T$_2$ grid. \\
		
		Valid support
		& Intracranial valid voxels
		& The HD-BET mask is intersected with finite, in-range, sum-to-one-valid tissue fractions. \\
		
		Excluded subjects
		& 16
		& Fifteen fail to meet the brain-mask criteria, and IXI260 is excluded because of registration failure. \\
		
		Independent evaluation masks
		& 99 test subjects
		& SynthSeg-derived CSF--GM and GM--WM interfaces are used for offline regional evaluation. \\
		\bottomrule
	\end{tabular}
	\endgroup
\end{table*}
\FloatBarrier

\subsection{Degradation Protocol and Implementation Details}

\noindent\textbf{Degradation protocol.}
LR inputs are synthesized by a frequency-domain truncation~\citep{downsample}. For a scale factor $s$, we retain the central $1/s \times 1/s$ region of k-space and set the remaining high-frequency coefficients to zero before inverse Fourier reconstruction. This retains 25\% and 6.25\% of k-space for 2$\times$ and 4$\times$, respectively.

\begin{figure*}[!tbp]
	\centering
	\includegraphics[width=0.98\textwidth]{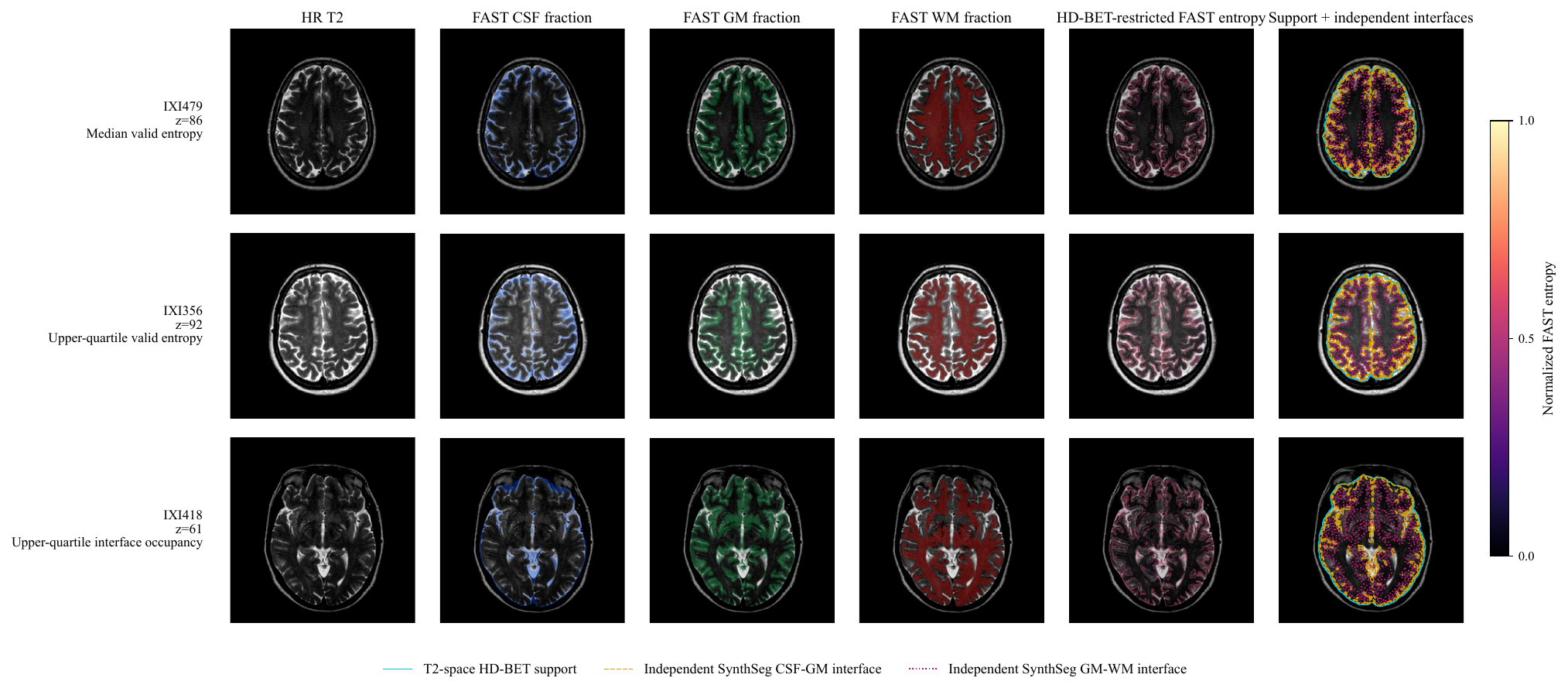}
	\caption{Qualitative validation of the IXI PVE pseudo-labels. FAST estimates CSF, GM, and WM fractions from brain-extracted T$_1$/T$_2$/PD images aligned in the T$_2$ reference space, and tissue-mixture entropy is shown only within the quality-controlled intracranial support. SynthSeg-derived CSF--GM and GM--WM interfaces serve as independent automated anatomical references. The three cases correspond to the median entropy, upper-quartile entropy, and upper-quartile interface occupancy within the eligible test cohort.}
	\label{fig:pve_qualitative_validation}
\end{figure*}

\noindent\textbf{Implementation Details.}
Our approach is implemented in PyTorch and trained on a single NVIDIA RTX 5090 GPU. We trained the model for 80 epochs using Adam Optimizer with a batch size of 4. The initial learning rate is $2\times10^{-4}$ and gradually reduced to $1\times10^{-6}$ using a cosine annealing schedule. For the IXI dataset, the training objective is $\mathcal{L}_{\mathrm{charb}} + \mathcal{L}_{\mathrm{PBR}}$, with $\alpha_{\mathrm{pve}}=0.1$. For the fastMRI dataset, AGW-backbone is trained from scratch using only $\mathcal{L}_{\mathrm{charb}}$.
Fig.~\ref{fig:pve_qualitative_validation} shows the qualitative validation layout for the IXI PVE pseudo-labels.

\subsection{Quality Control of the PVE Pseudo-label Sidecars}
\label{sec:pve_quality_control}

The PVE pipeline retains 475 of the 491 IXI subjects after brain-mask, registration, tissue-fraction, and sidecar quality control, including all 99 test subjects. Table~\ref{tab:pve_certification_summary} summarizes the cohort and validated sidecar resources. During training, FAST-derived tissue-mixture entropy defines the PBR weights; independent test-only SynthSeg segmentations define CSF--GM and GM--WM evaluation interfaces. Full construction and quality-control procedures are provided in Appendices A--D of the Supplementary Material.
Fig.~\ref{fig:pve_entropy_boundary_consistency} summarizes the spatial consistency analysis between FAST-derived tissue-mixture entropy and independent SynthSeg interfaces.

\begin{figure}[!tbp]
	\centering
	\includegraphics[width=\linewidth]{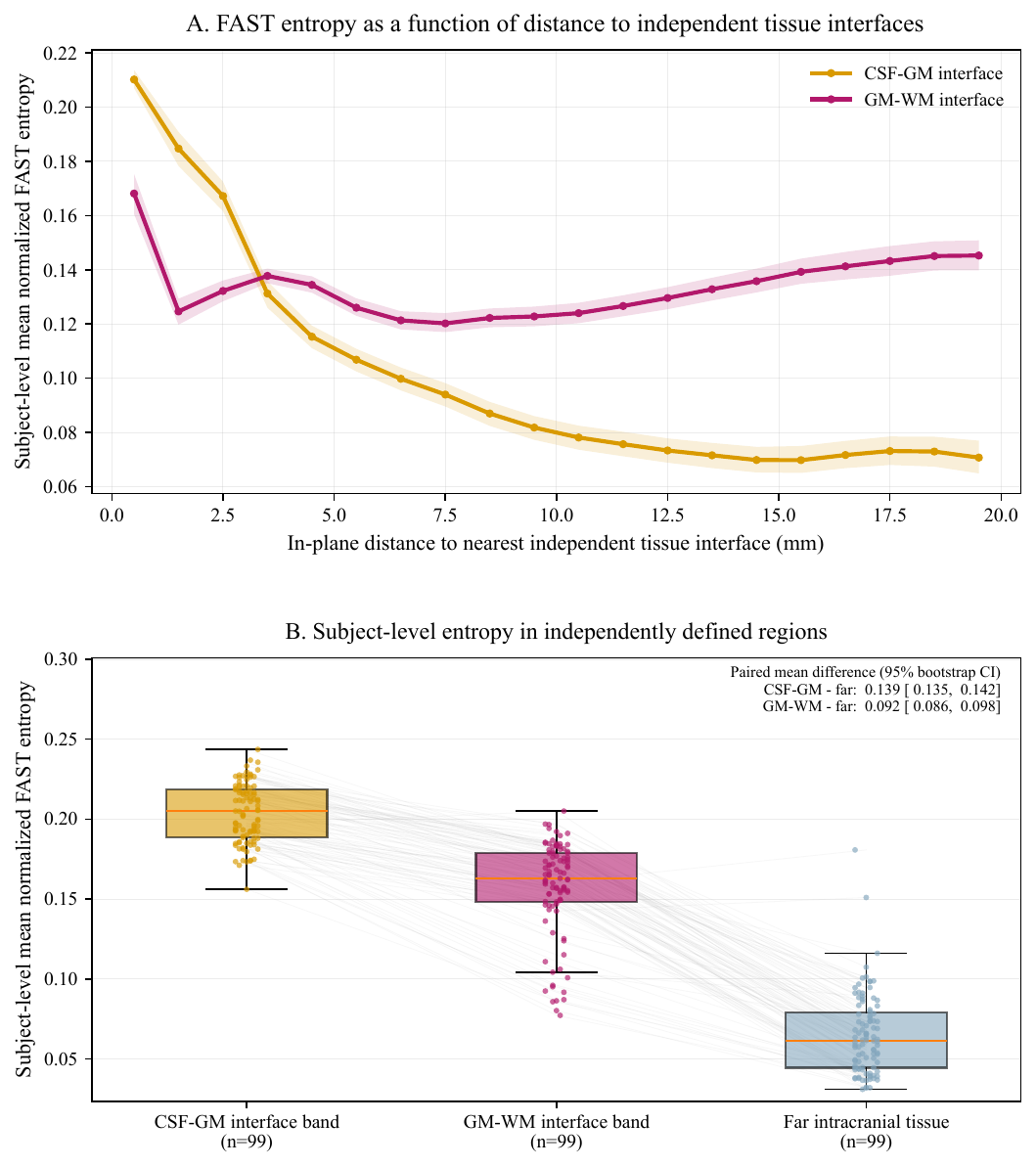}
	\caption{Spatial consistency between FAST-derived tissue-mixture entropy and independent SynthSeg tissue interfaces in 99 test subjects. Panel A shows subject-level mean entropy within the valid PVE support as a function of the in-plane distance to the CSF--GM and GM--WM interfaces; shaded regions indicate 95\% bootstrap confidence intervals. Panel B compares entropy within 2~mm of each interface with that in intracranial tissue at least 5~mm away. This analysis evaluates spatial agreement with an independent automated anatomical reference rather than absolute tissue-fraction accuracy.}
	\label{fig:pve_entropy_boundary_consistency}
\end{figure}
\label{sec:methods_metrics_statistics}

\noindent\textbf{Compared methods.}
We compare the evaluated configuration (AGW-PBR on IXI and AGW-backbone on fastMRI) with representative MRI SR baselines from several model families: SRCNN~\citep{srcnn}, RCAN~\citep{rcab}, SwinIR~\citep{swinir}, HAT~\citep{hat}, ArSSR~\citep{arssr}, NExpR~\citep{nexpr2025}, GaussianSR~\citep{gaussiansr}, and Res-SRDiff~\citep{ressrdiff}. To adapt the 3D-based ArSSR to our 2D in-plane MRI SR setting, we replace its 3D convolutional layers with their 2D counterparts. GaussianSR, NExpR, and Res-SRDiff retain their original reconstruction frameworks, with only the input and output layers adapted for single-channel MR images. Res-SRDiff follows its original configuration with 15 diffusion and sampling steps, while GaussianSR retains its $128\times128$ patch-based training strategy. All comparisons use publicly available codebases, and all models are retrained on our MRI datasets until convergence.

\noindent\textbf{Evaluation metrics and statistical analysis.}
We evaluate reconstruction using full-image PSNR and SSIM for the IXI comparison, fastMRI external evaluation, and ablation analysis. For the IXI $4\times$ experiment, regional PSNR and global masked-vector SSIM are additionally computed within independent SynthSeg-defined interface and non-interface intracranial regions, where interfaces are obtained by one three-dimensional six-neighbour dilation of the CSF--GM, GM--WM, and CSF--WM boundaries. Regional evaluation uses 10,872 eligible slices from all 99 test subjects, with slice-level scores averaged within each subject and then across seeds.

\begin{table*}[t]
	\centering
	\caption{Quantitative results on the IXI dataset at $2\times$ and $4\times$ upsampling. Slice-level scores are averaged within each subject, and results from seeds 42, 43, and 44 are then averaged within subject before reporting the cohort mean $\pm$ sample standard deviation. Bold: best. Underline: second-best.}
	\label{tab:ixi_main_quantitative}
	\small
	\setlength{\tabcolsep}{5pt}
	\begin{adjustbox}{max width=\textwidth}
		\begin{tabular}{lcccc}
			\toprule
			Method & $2\times$ PSNR $\uparrow$ & $2\times$ SSIM $\uparrow$ & $4\times$ PSNR $\uparrow$ & $4\times$ SSIM $\uparrow$ \\
			\midrule
			SRCNN & $36.130 \pm 1.076$ & $0.9711 \pm 0.0049$ & $29.157 \pm 0.851$ & $0.8972 \pm 0.0122$ \\
			RCAN & $39.498 \pm 1.085$ & $0.9827 \pm 0.0043$ & $31.194 \pm 0.822$ & $0.9280 \pm 0.0108$ \\
			SwinIR & $39.620 \pm 1.094$ & $0.9830 \pm 0.0042$ & $\underline{31.556 \pm 0.815}$ & $ 0.9314 \pm 0.0105$ \\
			HAT & $38.414 \pm 1.021$ & $0.9799 \pm 0.0044$ & $30.393 \pm 0.815$ & $0.9179 \pm 0.0113$ \\
			ArSSR & $39.638 \pm 1.093$ & $0.9831 \pm 0.0043$ & $31.383 \pm 0.829$ & $0.9313 \pm 0.0107$ \\
			GaussianSR & $39.626 \pm 1.095$ & $\underline{0.9833 \pm 0.0043}$ & $31.386 \pm 0.834$ & $0.9307 \pm 0.0109$ \\
			NExpR & $\underline{39.710 \pm 1.100}$ & $0.9829 \pm 0.0042$ & $31.406 \pm 0.827$ & $\underline{0.9326 \pm 0.0107}$ \\
			Res-SRDiff & $38.579 \pm 1.112$ & $0.9779 \pm 0.0052$ & $30.568 \pm 0.885$ & $0.9190 \pm 0.0136$ \\
			\textbf{AGW-PBR (Ours)} & $\mathbf{39.881 \pm 1.130}$ & $\mathbf{0.9834 \pm 0.0043}$ & $\mathbf{32.844 \pm 1.008}$ & $\mathbf{0.9460 \pm 0.0112}$ \\
			\bottomrule
		\end{tabular}
	\end{adjustbox}
\end{table*}

\subsection{Results on the IXI Dataset}
\label{sec:ixi_results}

\begin{figure*}[t]
	\centering
	\includegraphics[width=0.98\textwidth]{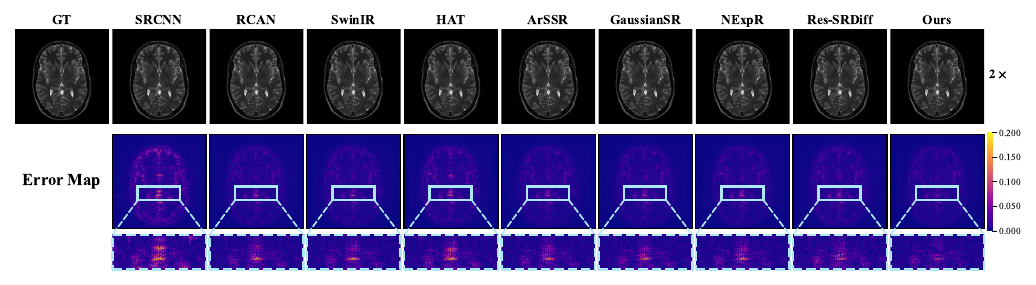}
	\vspace{2pt}
	\includegraphics[width=0.98\textwidth]{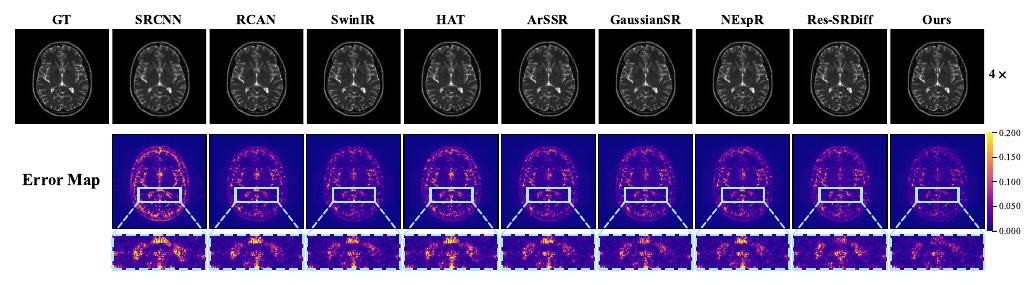}
	\caption{Qualitative results on the IXI dataset under $2\times$ and $4\times$ upsampling. Blue boxes mark the ROI in the error maps, where stronger residual intensity indicates larger deviation from the HR reference.}
	\label{fig:ixi_comparison_multiscale}
\end{figure*}

\noindent\textbf{Qualitative Results.} Fig.~\ref{fig:ixi_comparison_multiscale} compares representative reconstructions and absolute-error maps at $2\times$ and $4\times$. Differences are more pronounced at 4$\times$, particularly around ventricular and tissue-interface regions. AGW-PBR shows lower residual intensity in the highlighted ROIs while preserving local anatomical continuity.

\noindent\textbf{Quantitative Results.}
Tables~\ref{tab:ixi_main_quantitative} and~\ref{tab:ixi_pve_regional_4x} summarize the IXI results. AGW-PBR achieves the highest PSNR and SSIM at both upsampling factors.

At $4\times$, it also performs best in both SynthSeg-defined interface and non-interface intracranial regions, showing improved tissue-interface reconstruction without sacrificing overall intracranial fidelity.

\begin{table*}[t]
	\centering
	\caption{Independent SynthSeg regional reconstruction fidelity on the IXI dataset at $4\times$. Every method and seed uses the same canonical manifest of 10,872 eligible slices from 99 test subjects. Slice-level metrics are averaged within subject, and seeds 42, 43, and 44 are then averaged within subject before reporting the cohort mean $\pm$ sample standard deviation. Interface regions are derived from SynthSeg labels on the locked HR T$_2$ images, and regional SSIM uses the global masked-vector statistic. Bold: best. Underline: second-best.}
	\label{tab:ixi_pve_regional_4x}
	\setlength{\tabcolsep}{3.5pt}
	\resizebox{\textwidth}{!}{%
		\begin{tabular}{lcccc}
			\toprule
			Method & Combined-interface PSNR $\uparrow$ & Combined-interface SSIM $\uparrow$ & Non-interface PSNR $\uparrow$ & Non-interface SSIM $\uparrow$ \\
			\midrule
			SRCNN & $23.705 \pm 1.066$ & $0.8977 \pm 0.0159$ & $25.408 \pm 0.918$ & $0.9183 \pm 0.0143$ \\
			RCAN & $25.852 \pm 0.987$ & $0.9379 \pm 0.0104$ & $27.165 \pm 0.891$ & $0.9439 \pm 0.0114$ \\
			SwinIR & $26.128 \pm 0.978$ & $\underline{0.9440 \pm 0.0096}$ & $27.490 \pm 0.910$ & $\underline{0.9484 \pm 0.0110}$ \\
			HAT & $25.023 \pm 0.984$ & $0.9261 \pm 0.0118$ & $26.376 \pm 0.871$ & $0.9338 \pm 0.0125$ \\
			ArSSR & $26.090 \pm 0.981$ & $0.9413 \pm 0.0097$ & $27.403 \pm 0.899$ & $0.9471 \pm 0.0110$ \\
			GaussianSR & $26.157 \pm 0.989$ & $0.9421 \pm 0.0097$ & $\underline{27.531 \pm 0.902}$ & $0.9481 \pm 0.0109$ \\
			NExpR & $\underline{26.323 \pm 0.987}$ & $0.9417 \pm 0.0097$ & $27.418 \pm 0.899$ & $0.9469 \pm 0.0110$ \\
			Res-SRDiff & $25.449 \pm 1.053$ & $0.9333 \pm 0.0112$ & $26.667 \pm 0.972$ & $0.9386 \pm 0.0136$ \\
			\textbf{AGW-PBR (Ours)} & $\mathbf{27.784 \pm 1.177}$ & $\mathbf{0.9590 \pm 0.0096}$ & $\mathbf{28.909 \pm 1.129}$ & $\mathbf{0.9623 \pm 0.0097}$ \\
			\bottomrule
		\end{tabular}%
	}
\end{table*}

\noindent\textbf{Tissue-transition fidelity.}
We further assess structural fidelity at the independently defined CSF--GM and GM--WM tissue transitions under the IXI 4$\times$ setting. 
Table~\ref{tab:ixi_transition_fidelity} reports the normalized gradient-vector reconstruction error on the 99-subject test cohort. AGW-PBR achieves the lowest error at both interfaces, showing that the regional reconstruction gains extend to local tissue-transition gradients.

Boundary-profile RMSE provides complementary evidence on HR-eligible profiles, with AGW-PBR achieving the lowest mean error at both interfaces. The complete profile definition, cohort coverage, and results are provided in Appendix~E.
\begin{table}[t]
	\centering
	\caption{Tissue-transition gradient fidelity on the IXI dataset at $4\times$ (seed 42).
		Lower values indicate closer agreement with the HR gradient field.
		Bold: best. Underline: second-best.}
	\label{tab:ixi_transition_fidelity}
	
	\small
	\setlength{\tabcolsep}{6pt}
	
	\begin{adjustbox}{max width=\linewidth}
		\begin{tabular}{lcc}
			\toprule
			Method
			& CSF--GM Error $\downarrow$
			& GM--WM Error $\downarrow$ \\
			\midrule
			SRCNN
			& $0.6609 \pm 0.0304$
			& $0.7617 \pm 0.0416$ \\
			
			RCAN
			& $0.5142 \pm 0.0272$
			& $0.6233 \pm 0.0369$ \\
			
			SwinIR
			& $0.4951 \pm 0.0268$
			& $\underline{0.5950 \pm 0.0399}$ \\
			
			HAT
			& $0.5595 \pm 0.0276$
			& $0.6628 \pm 0.0344$ \\
			
			ArSSR
			& $0.4967 \pm 0.0270$
			& $0.6018 \pm 0.0391$ \\
			
			GaussianSR
			& $0.4949 \pm 0.0264$
			& $0.6014 \pm 0.0386$ \\
			
			NExpR
			& $\underline{0.4866 \pm 0.0265}$
			& $0.6012 \pm 0.0393$ \\
			
			Res-SRDiff
			& $0.5435 \pm 0.0333$
			& $0.6579 \pm 0.0544$ \\
			
			\textbf{AGW-PBR (Ours)}
			& $\mathbf{0.4136 \pm 0.0367}$
			& $\mathbf{0.5148 \pm 0.0516}$ \\
			\bottomrule
		\end{tabular}
	\end{adjustbox}
\end{table}

\subsection{Results on the fastMRI Dataset}
\label{sec:fastmri_results}
\begin{figure*}[!tbp]
	\centering
	\includegraphics[width=0.98\textwidth]{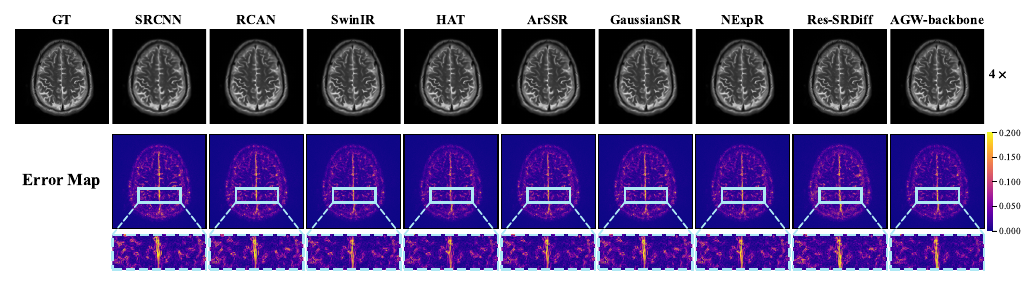}
	\caption{Qualitative results on the fastMRI dataset under $4\times$ upsampling. Blue boxes mark the ROI in the error maps, where stronger residual intensity indicates larger deviation from the HR reference.}
	\label{fig:fastmri_comparison_4x}
\end{figure*}

\noindent\textbf{Qualitative Results.}
Fig.~\ref{fig:fastmri_comparison_4x} shows representative $4\times$ fastMRI results. AGW-backbone exhibits lower error-map intensity in the highlighted ROIs, indicating cleaner local reconstruction. 

\noindent\textbf{Quantitative Results.}
Table~\ref{tab:fastmri_4x_main} reports the PVE-free fastMRI evaluation under full-image supervision. AGW-backbone achieves the highest PSNR and SSIM, demonstrating strong external-dataset performance without PVE-based weighting or tissue-interface supervision.

\begin{table}[t]
	\centering
	\caption{Quantitative results on fastMRI at $4\times$. Values are subject-level mean $\pm$ standard deviation. Bold: best. Underline: second-best.}
	\label{tab:fastmri_4x_main}
	
	\small
	\setlength{\tabcolsep}{6pt}
	
	\begin{adjustbox}{max width=\linewidth}
		\begin{tabular}{lcc}
			\toprule
			Method & PSNR $\uparrow$ & SSIM $\uparrow$ \\
			\midrule
			SRCNN & $31.6443 \pm 1.0451$ & $0.9166 \pm 0.0344$ \\
			RCAN & $32.3261 \pm 1.0264$ & $0.9240 \pm 0.0347$ \\
			SwinIR & $33.3219 \pm 1.0505$ & $\underline{0.9331 \pm 0.0351}$ \\
			HAT & $33.0450 \pm 1.0336$ & $0.9319 \pm 0.0347$ \\
			ArSSR & $32.9520 \pm 1.0275$ & $0.9299 \pm 0.0349$ \\
			GaussianSR & $32.5919 \pm 1.0271$ & $0.9267 \pm 0.0347$ \\
			NExpR & $\underline{33.4477 \pm 1.0544}$ & $0.9327 \pm 0.0350$ \\
			Res-SRDiff & $31.8944 \pm 0.9652$ & $0.9109 \pm 0.0391$ \\
			\textbf{AGW-backbone}
			& $\mathbf{33.6827 \pm 1.0681}$
			& $\mathbf{0.9448 \pm 0.0347}$ \\
			\bottomrule
		\end{tabular}
	\end{adjustbox}
\end{table}


\subsection{Ablation Study}
\label{sec:ablation}

The ablation study examines valid-support reconstruction, entropy-weight correspondence, and dictionary assignment. Full-image PSNR/SSIM are evaluated over 12,870 slices from 99 subjects, while SynthSeg interface metrics use a fixed set of 10,872 eligible slices. All slice-level metrics are averaged per subject before aggregation.

\noindent\textbf{Ablation variants.}
A0 is the backbone trained with only full-image Charbonnier loss; A1 adds uniform valid-support reconstruction; and A2 is the complete AGW-PBR with PVE-entropy weighting. A3 and A4 replace the validated entropy field with spatially shuffled and deterministic random fields, respectively. A5 retains PVE-entropy weighting but replaces soft dictionary assignment with the Hard-ST rule defined below.

\begin{table*}[t]
	\centering
	\caption{Ablation study on T$_2$-weighted IXI images at $4\times$ using seed 42. Full-image metrics are computed over all 12,870 test slices. SynthSeg interface metrics use the canonical set of 10,872 eligible slices within the independent combined tissue-interface band. Slice-level values are averaged within each subject, and the table reports the mean $\pm$ standard deviation across 99 subjects. Bold: best. Underline: second-best. AGW denotes Anatomy-Guided Gaussian-Parameter Warping, and PBR denotes PVE-Balanced Reconstruction.}
	\label{tab:ablation_ixi4x_seed42}
	\resizebox{\textwidth}{!}{
		\begin{tabular}{llccccccc}
			\toprule
			ID & Variant & Assignment & Valid-support term & Weighting field
			& PSNR $\uparrow$ & SSIM $\uparrow$
			& SynthSeg interface PSNR $\uparrow$ & SynthSeg interface SSIM $\uparrow$ \\
			\midrule
			A0 & AGW-backbone
			& Soft & -- & --
			& 31.8351 $\pm$ 0.8771
			& 0.9364 $\pm$ 0.0105
			& 26.7469 $\pm$ 1.0569
			& 0.9491 $\pm$ 0.0095 \\
			
			A1 & Uniform-support control
			& Soft & \checkmark & Uniform on valid support
			& \underline{32.1346 $\pm$ 0.8929}
			& 0.9397 $\pm$ 0.0107
			& \underline{26.8642 $\pm$ 1.0726}
			& \underline{0.9511 $\pm$ 0.0088} \\
			
			A2 & AGW-PBR
			& Soft & \checkmark & Validated PVE entropy
			& \textbf{32.8897 $\pm$ 0.9844}
			& \textbf{0.9463 $\pm$ 0.0111}
			& \textbf{27.7167 $\pm$ 1.1519}
			& \textbf{0.9585 $\pm$ 0.0094} \\
			
			A3 & PBR-shuffled
			& Soft & \checkmark & Shuffled PVE entropy
			& 32.0074 $\pm$ 0.8976
			& \underline{0.9400 $\pm$ 0.0105}
			& 26.8137 $\pm$ 1.0560
			& 0.9497 $\pm$ 0.0095 \\
			
			A4 & PBR-random
			& Soft & \checkmark & Deterministic random field
			& 31.8577 $\pm$ 0.8839
			& 0.9371 $\pm$ 0.0105
			& 26.8094 $\pm$ 1.0698
			& 0.9494 $\pm$ 0.0094 \\
			
			A5 & Hard-ST + PBR
			& Hard-ST & \checkmark & Validated PVE entropy
			& 31.8290 $\pm$ 0.8889
			& 0.9340 $\pm$ 0.0104
			& 26.7764 $\pm$ 1.0726
			& 0.9492 $\pm$ 0.0094\\
			\bottomrule
		\end{tabular}
	}
\end{table*}

\noindent\textbf{Effect of valid-support reconstruction and entropy modulation.}
A0, A1, and A2 isolate the effects of valid-support reconstruction and entropy modulation:
\begin{equation}
	\mathcal L_{\mathrm{A0}} = \mathcal L_{\mathrm{charb}},
	\ 
	\mathcal L_{\mathrm{A1}} = \mathcal L_{\mathrm{charb}} + \mathcal L_{\mathrm{uniform}},
	\ 
	\mathcal L_{\mathrm{A2}} = \mathcal L_{\mathrm{charb}} + \mathcal L_{\mathrm{PBR}},
	\label{eq:ablation_a0_a1_a2}
\end{equation} 
where
\begin{equation}
	\mathcal L_{\mathrm{uniform}}
	=
	\left.\mathcal L_{\mathrm{PBR}}\right|_{\alpha_{\mathrm{pve}}=0}.
	\label{eq:ablation_uniform}
\end{equation}
Thus, A1 applies uniform weighting within the valid support, whereas A2 uses PVE entropy for spatial modulation.

As shown in Table~\ref{tab:ablation_ixi4x_seed42}, A1 improves both full-image and interface reconstruction over A0, while A2 further improves all metrics, with larger gains at tissue interfaces. This supports the contributions of valid-support supervision and entropy-based emphasis on anatomical transitions.

\noindent\textbf{Effect of anatomical alignment in PVE weighting.}
A3 and A4 test whether PVE weighting benefits from anatomical alignment or merely nonuniform spatial weights. For the $n$-th subject--slice pair, let
\(\mathcal V_n\) denote its valid PVE support. A3 and A4 replace the validated guide with
\begin{align}
	G_{\mathrm{A3}}^{(n)}(\mathbf p)
	&=
	H_{\mathrm{PVE}}^{(n)}
	\!\left(
	\rho_n(\mathbf p)
	\right),
	\qquad
	\rho_n:\mathcal V_n\rightarrow\mathcal V_n,
	\label{eq:ablation_shuffled}
	\\
	G_{\mathrm{A4}}^{(n)}(\mathbf p)
	&=
	R^{(n)}(\mathbf p),
	\qquad
	R^{(n)}(\mathbf p)\sim\mathcal U[0,1),
	\qquad
	\mathbf p\in\mathcal V_n,
	\label{eq:ablation_random}
\end{align}
respectively. A3 spatially shuffles the entropy field and A4 uses a fixed random field within the valid support. Both follow the same weighting and PBR formulation as A2. Their consistently lower full-image and interface metrics indicate that A2 benefits from the anatomical correspondence of PVE entropy rather than nonuniform weighting alone.

\noindent\textbf{Effect of the assignment rule.}
A5 replaces the soft dictionary assignment in A2 with one-hot top-$1$ routing while retaining validated PVE-entropy weighting. Let \(W_k^{\mathrm{soft}}(\mathbf p)\) denote the soft assignment in Eq.~\eqref{eq:soft_assignment}. The hard assignment used in the forward pass and its straight-through form are defined as
\begin{align}
	\widehat{W}_k(\mathbf p)
	&=
	\mathbb{I}
	\!\left[
	k=
	\underset{1\leq j\leq K}{\arg\max}\;
	W_j^{\mathrm{soft}}(\mathbf p)
	\right],
	\label{eq:ablation_a5_onehot}
	\\
	W_k^{\mathrm{ST}}(\mathbf p)
	&=
	\operatorname{sg}
	\!\left(
	\widehat{W}_k(\mathbf p)
	\right)
	-
	\operatorname{sg}
	\!\left(
	W_k^{\mathrm{soft}}(\mathbf p)
	\right)
	+
	W_k^{\mathrm{soft}}(\mathbf p),
	\label{eq:ablation_a5_st}
\end{align}
where \(\mathbb{I}[\cdot]\) is the indicator function and \(\operatorname{sg}(\cdot)\) denotes the stop-gradient operator. The forward pass uses the one-hot assignment, while gradients are propagated through the soft assignment. The resulting dictionary
\begin{equation}
	\boldsymbol{\eta}_{\mathrm{base}}^{\mathrm{A5}}(\mathbf p)
	=
	\sum_{k=1}^{K}
	W_k^{\mathrm{ST}}(\mathbf p)\,
	\mathbf b_k .
	\label{eq:ablation_a5_aggregation}
\end{equation}

A5 performs near A0 and remains below A2 on all four metrics. Since A2 and A5 differ only in the assignment rule, the results support soft assignment, which allows each spatial location to combine multiple latent basis vectors instead of selecting a single basis.

\subsection{Application on Brain Tumor Segmentation Task}

We evaluate downstream brain tumor segmentation on BraTS2023\footnote{\url{https://www.med.upenn.edu/cbica/brats}}. T2 images are degraded at $4\times$ and reconstructed using IXI-trained (seed 42) SwinIR, NExpR, GaussianSR, AGW-backbone, and AGW-PBR checkpoints without fine-tuning. A frozen segmentation probe trained on native HR BraTS T2 images evaluates whole-tumor (WT) segmentation using Dice and HD95 on 240 test subjects.

\begin{table}[t]
	\centering
	\caption{Downstream whole-tumor segmentation results on the BraTS test cohort under $4\times$ degradation. Values are subject-level mean $\pm$ standard deviation. The SR models use the checkpoints trained on IXI with seed 42 and are transferred to BraTS without fine-tuning. Bold: best. Underline: second-best.}
	\label{tab:brats_segmentation}
	\begin{tabular}{lcc}
		\toprule
		Input / Method & WT Dice $\uparrow$ & WT HD95 (mm) $\downarrow$ \\
		\midrule
		HR             & 0.8371 $\pm$ 0.1302 & 15.689 $\pm$ 18.955 \\
		Canonical LR   & 0.4512 $\pm$ 0.1928 & 70.929 $\pm$ 14.693 \\
		\midrule
		SwinIR         & 0.7020 $\pm$ 0.1822 & 57.755 $\pm$ 18.979 \\
		NExpR          & \underline{0.7210 $\pm$ 0.1748} & 53.843 $\pm$ 20.958 \\
		GaussianSR     & \textbf{0.7234 $\pm$ 0.1733} & 52.530 $\pm$ 21.026 \\
		AGW-backbone   & 0.7055 $\pm$ 0.1824 & \underline{52.507 $\pm$ 20.525} \\
		AGW-PBR (Ours)        & 0.7158 $\pm$ 0.1785 & \textbf{49.881 $\pm$ 22.519} \\
		\bottomrule
	\end{tabular}
\end{table}

\begin{figure*}[t]
	\centering
	\includegraphics[width=\textwidth]{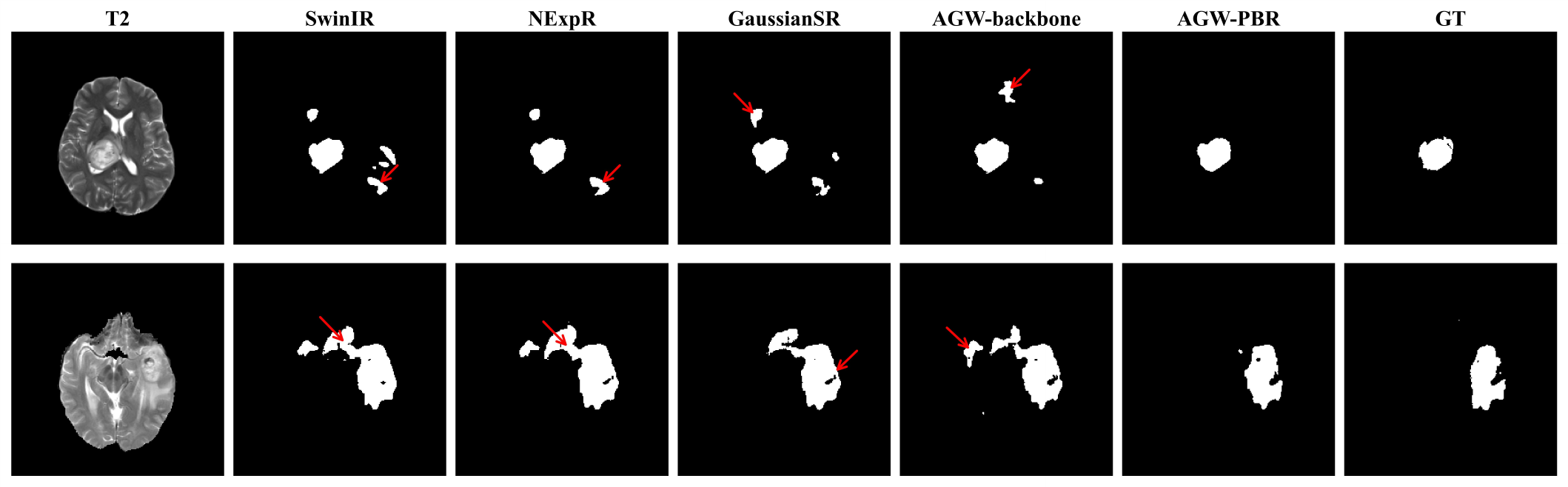}
	\caption{Qualitative comparison of downstream whole-tumor segmentation on BraTS under $4\times$ degradation. 
		The first column shows the corresponding T2 image, the intermediate columns show segmentation predictions obtained from different SR reconstructions, and the last column shows the ground-truth WT mask. 
		Red arrows mark representative spurious or disconnected predictions. 
		}
	\label{fig:brats_segmentation}
\end{figure*}

As shown in Table~\ref{tab:brats_segmentation}, AGW-PBR achieves competitive WT Dice and the lowest HD95 among the evaluated SR methods, indicating stronger preservation of tumor-boundary information after reconstruction. The qualitative examples in Fig.~\ref{fig:brats_segmentation} are consistent with the quantitative results. Several competing reconstructions introduce isolated responses or deviations around the tumor region, whereas AGW-PBR yields more spatially coherent WT predictions. The PVE-balanced model also improves both downstream metrics over AGW-backbone, showing that the benefit of PBR extends to pathological images that were unseen during SR training.

\captionsetup{justification=raggedright}

%% file: sections/conclusion.tex
\section{Discussion}
\label{sec:discussion}

AGW-PBR improves full-image reconstruction across the evaluated IXI scales and retains this advantage within independently defined interface and non-interface regions. The $4\times$ structural analysis further shows lower normalized gradient-vector error at both CSF--GM and GM--WM interfaces. Boundary-profile RMSE provides consistent supporting evidence on the smaller HR-eligible profile populations. These results connect the global reconstruction gains with fidelity at anatomically defined tissue transitions.

The ablation results support three components of the proposed formulation. Valid-support reconstruction improves the AGW backbone, entropy modulation provides an additional gain under the same support and loss coefficient, and shuffled/random controls show the importance of spatial correspondence. The Hard-ST comparison further favors soft latent-basis aggregation.

The external experiments address complementary generalization settings. The fastMRI experiment evaluates the independently trained LR-only backbone without PVE supervision or IXI initialization. The BraTS2023 experiment evaluates direct transfer of IXI-trained SR checkpoints through a frozen T$_2$-only tumor-segmentation probe and measures task-relevant information retention under this fixed downstream protocol.

The study has several limitations. The LR observations are generated through synthetic central $k$-space truncation, which represents only part of the noise, motion, reconstruction filtering, and acquisition variability encountered in clinical imaging. The model processes two-dimensional slices, leaving through-plane consistency and volumetric reconstruction untested. The study does not include paired clinical LR/HR acquisitions or reader assessments. The BraTS evaluation uses a single frozen downstream probe, while fastMRI evaluates only the PVE-free backbone. Future work should examine real LR/HR acquisitions, 3-D reconstruction, broader downstream tasks, reader studies, and independent scanners and cohorts.

\section{Conclusion}

AGW-PBR combines LR-only anatomy-guided warping with a training-time PVE-balanced objective derived from quality-controlled tissue-mixture entropy. Across the evaluated IXI settings, it improves full-image and regional reconstruction fidelity and achieves the lowest normalized gradient-vector error at both CSF--GM and GM--WM interfaces. The ablation results support valid-support reconstruction, spatially aligned entropy weighting, and soft latent assignment.

The PVE-free AGW backbone remains strong under independent fastMRI training, and the BraTS2023 frozen-probe transfer extends the evaluation to downstream tumor segmentation. Overall, the results support PVE-balanced reconstruction for brain MRI SR while preserving LR-only inference.